%% file: main.tex
\documentclass[10pt,twocolumn,letterpaper]{article}

\usepackage[pagenumbers]{cvpr}      
\usepackage{pifont} 

\input{preamble}

\usepackage{wrapfig}

\definecolor{cvprblue}{rgb}{0.21,0.49,0.74}
\usepackage[pagebackref,breaklinks,colorlinks,citecolor=cvprblue]{hyperref}

\title{Adaptive Illumination-Invariant Synergistic Feature Integration in a Stratified Granular Framework for Visible-Infrared Re-Identification}

\author{
Yuheng Jia \\
Oxford e-Research Centre \\
Department of Engineering Science \\
University of Oxford \\
Oxford, UK \\
{\tt\small yuheng.jia@eng.ox.ac.uk}
\and
Wesley Armour \\
Oxford e-Research Centre \\
Department of Engineering Science \\
University of Oxford \\
Oxford, UK \\
{\tt\small wes.armour@oerc.ox.ac.uk}
}

\begin{document}
\maketitle
\input{0_abstract}            
\input{1_intro}               
\input{2_lit_review}          
\input{3_proposal}            
\input{4_experiments}         

\input{5_conclusion}          

\clearpage  
{
    \small
    \bibliographystyle{ieeenat_fullname}
    \bibliography{main}
}


\end{document}

%% file: preamble.tex
\usepackage[dvipsnames]{xcolor}


%% file: 0_abstract.tex
\begin{abstract}

Visible-Infrared Person Re-Identification (VI-ReID) plays a crucial role in applications such as search and rescue, infrastructure protection, and nighttime surveillance. However, it faces significant challenges due to modality discrepancies, varying illumination, and frequent occlusions. To overcome these obstacles, we propose \textbf{AMINet}, an Adaptive Modality Interaction Network. AMINet employs multi-granularity feature extraction to capture comprehensive identity attributes from both full-body and upper-body images, improving robustness against occlusions and background clutter. The model integrates an interactive feature fusion strategy for deep intra-modal and cross-modal alignment, enhancing generalization and effectively bridging the RGB-IR modality gap. Furthermore, AMINet utilizes phase congruency for robust, illumination-invariant feature extraction and incorporates an adaptive multi-scale kernel MMD to align feature distributions across varying scales. Extensive experiments on benchmark datasets demonstrate the effectiveness of our approach, achieving a Rank-1 accuracy of $74.75\%$ on SYSU-MM01, surpassing the baseline by $7.93\%$ and outperforming the current state-of-the-art by $3.95\%$.

\end{abstract}

%% file: 1_intro.tex
\section{Introduction}

Visible-Infrared Person Re-identification (VI-ReID) is critical for modern security and surveillance, enabling the matching of individuals across spectral modalities in low-light conditions. Applications like nighttime surveillance, search and rescue, and infrastructure protection rely on IR imaging to complement RGB, capturing thermal radiation where visible light fails \cite{example30, example44, example28}. However, the modality gap between RGB and IR images presents significant challenges \cite{example37, example26, example55, example46}. RGB relies on visible light reflection, while IR captures thermal emission, leading to substantial differences in appearance, texture, and color \cite{example43, example29}. These discrepancies complicate the extraction of robust, modality-invariant features, and real-world factors like variable illumination, occlusions, and background clutter further add to the complexity of reliable VI-ReID \cite{example11, example48, example35, example56}.

To address the modality gap in Visible-Infrared Person Re-Identification (VI-ReID), several methods have been proposed. Generative Adversarial Networks (GANs) are commonly used to generate consistent RGB-IR images, but they often suffer from artifacts that degrade image quality and impair recognition \cite{example18, example30, example47}. Shared-specific feature transfer methods, like Cross-Modal Shared-Specific Feature Transfer (cm-SSFT), integrate shared and modality-specific features for better cross-modal fusion \cite{example24, example50}. Hybrid Dual-Path Networks enhance alignment by leveraging shared and specific parameter layers with consistency constraints \cite{example55, example40, example33, example56}. Graph Neural Networks (GNNs) model complex inter-modal relationships, using node-level information propagation to strengthen feature representation \cite{example31, example46}. Semantic enhancements, including human pose estimation and attribute integration, improve discriminative capability \cite{example43, example35}. Lastly, domain adaptation techniques such as adversarial training align features across modalities but still face challenges in effectively capturing identity-specific details \cite{example26}.

In this paper, we propose a novel framework to tackle the complex challenges of Visible-Infrared Person Re-Identification (VI-ReID), emphasizing comprehensive multi-granularity feature extraction and robust cross-modality alignment. Our framework, the Hierarchical Multi-Granular Dual-Branch Network (HMG-DBNet), utilizes a dual-branch architecture to independently process full-body and upper-body images. This design captures both high-level semantic information and detailed fine-grained features, enabling the model to effectively learn distinctive appearance attributes even under partial occlusion. By integrating representations from both full-body and upper-body inputs, HMG-DBNet extracts comprehensive global features and critical upper-body details, such as shoulder posture and clothing texture, enhancing its capacity to differentiate individual identities.

To effectively bridge the modality gap between RGB and infrared (IR) images, we introduce the Interactive Feature Fusion Strategy (IFFS). ICMIAS addresses the common issue of fragmented feature representation by integrating intra-modality fusion with cross-modality alignment, ensuring a cohesive combination of global and local features while simultaneously aligning data across RGB and IR modalities. The strategy leverages a nonlinear fusion mechanism, enhancing the interaction between modalities and facilitating the learning of robust, modality-invariant representations. To further mitigate challenges posed by lighting variations, we incorporate a phase congruency-based feature extraction method, Phase-Enhanced Structural Attention Module (PESAM). This approach captures illumination-invariant structural features, providing a consistent foundation for effective cross-modality alignment. Complementing this, the PESAM selectively focuses on salient structural regions, refining the alignment process by prioritizing critical features. Together, these innovations enhance the model's ability to achieve precise feature alignment and robust generalization, offering a comprehensive solution for cross-modality person re-identification tasks.

Traditional Maximum Mean Discrepancy (MMD) methods rely on a fixed single-kernel, limiting their adaptability to complex, multi-scale data patterns. To address this, we introduce the Adaptive Multi-Scale Kernel MMD (AMK-MMD) module, which uses multi-scale Gaussian kernels with adaptive bandwidth selection and learnable weights for greater flexibility in feature alignment. To improve scalability, we also optimize computational efficiency through mini-batch processing and vectorized computations, reducing complexity and making AMK-MMD suitable for large datasets. These integrated innovations achieve precise feature alignment and robust generalization, providing a scalable and state-of-the-art solution for real-world cross-modality person re-identification.

The main contributions of this paper are summarized as follows:

\begin{itemize}
    \item HMG-DBNet processes full-body and half-body images separately, capturing complementary global and detailed features for better RGB-IR alignment and re-identification robustness.

    \item IFFS integrates intra-modality fusion with cross-modality alignment, reducing feature-level discrepancies and enhancing recognition accuracy by combining global and local features.

    \item PESAM leverages phase congruency for illumination-invariant feature extraction and uses edge-guided attention to focus on key structural regions for precise alignment.

    \item AMK-MMD employs multi-scale kernel fusion with adaptive bandwidth selection, capturing fine-grained discrepancies and improving alignment efficiency for scalable cross-modality tasks.
    
\end{itemize}

%% file: 2_lit_review.tex
\section{Related Work}

\vspace{1em} 
\noindent\textbf{Cross-Modal Person Re-Identification.} Cross-modal Re-ID (VI-ReID and VT Re-ID) tackles the challenge of matching identities between visible and infrared images. Supervised approaches, such as DTRM and CAJ, align modality-independent features at both feature and pixel levels, achieving strong cross-modality performance \cite{example23, example24}. Unsupervised methods, including Cluster Contrast, ICE, ADCA, and PGM, utilize clustering and graph-based strategies to establish cross-modal correspondence without relying on labeled data \cite{example26, example27, example28, example29}. GAN-based models, like Align-GAN, enhance alignment by fusing features and performing pixel-level transformations. Auxiliary techniques, such as X-modality generation and grayscale transformation, further reduce the modality gap, improving feature consistency across domains \cite{example30, example31, example32, example33}. Despite their effectiveness, these methods often face scalability issues, high annotation costs, and sensitivity to noise, limiting their applicability in large-scale settings \cite{example36, example37, example38, example39}.

\noindent\textbf{Data Augmentation.} Data augmentation is crucial for mitigating modality discrepancies and enhancing sample diversity in person re-identification. Traditional methods include modality transformation, color space conversion, and noise addition, which help improve generalization \cite{example57, example58}. Advanced techniques like Modality Mixed Augmentation (M-CutMix) create mixed samples to facilitate robust feature learning across visible and infrared modalities \cite{example60, example61}. Semantic-guided pixel sampling selectively swaps clothing regions (e.g., shirts, pants) using human parsing models, generating structurally consistent samples for better generalization \cite{example59, example64}. Inter-modality transformations and GAN-based synthetic generation further expand datasets by converting visible to infrared images, boosting cross-modality alignment \cite{example62, example63}. Overall, these augmentation strategies significantly enhance model performance by increasing sample diversity and addressing modality gaps.

%% file: 3_proposal.tex
\section{Proposed Method}

In cross-modality person re-identification, aligning RGB and IR features presents significant challenges due to differences in illumination and spectral characteristics. To address this, we propose a \textbf{Hierarchical Multi-Granular Dual-Branch Network (HMG-DBNet)}, which systematically captures comprehensive identity features by extracting and fusing multi-granularity information from both global and part-based views.

\begin{figure*}
\centering
    \includegraphics[width=\textwidth]{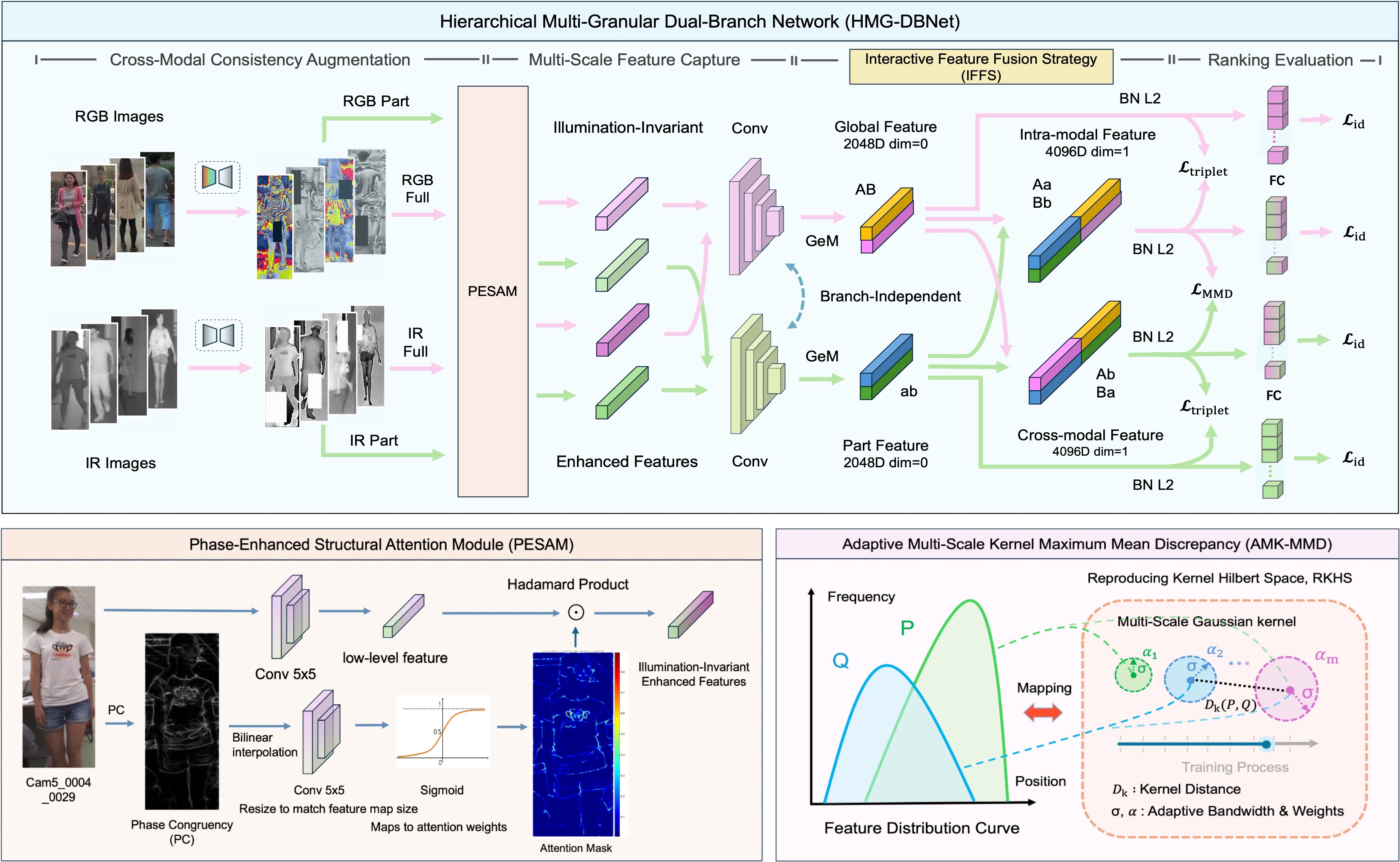}  
    \captionsetup{justification=justified, singlelinecheck=false}  
    \caption{The proposed framework employs an Interactive Feature Fusion Strategy(IFFS) within a dual-stream network (HMG-DBNet) to capture full-body and partial-body features effectively. PESAM utilizes phase congruency and edge-guided attention for robust RGB-IR alignment. AMK-MMD adaptively aligns feature distributions to reduce modality discrepancies. The multi-branch design, supervised by ID, MMD and Triplet losses, enhances the overall accuracy of cross-modality person re-identification.}
    \label{fig:query_results}
\end{figure*}

\subsection{Multi-Granular Feature Extraction and Fusion}

HMG-DBNet leverages two branches to handle full-body and half-body images in both RGB and IR modalities, enabling multi-scale identity feature extraction. Let \( X_{\text{rgb}}^{\text{global}} \in \mathbb{R}^{H \times W \times C} \) and \( X_{\text{ir}}^{\text{global}} \in \mathbb{R}^{H \times W \times C} \) represent full-body RGB and IR images, while \( X_{\text{rgb}}^{\text{part}} \in \mathbb{R}^{\frac{H}{2} \times W \times C} \) and \( X_{\text{ir}}^{\text{part}} \in \mathbb{R}^{\frac{H}{2} \times W \times C} \) are half-body images. The branches extract global and part-based features as:
\begin{equation}
    F_{\text{global}}^{m} = \phi_{\text{global}}(X_{\text{rgb}}^{\text{global}}), \quad F_{\text{global}}^{i} = \phi_{\text{global}}(X_{\text{ir}}^{\text{global}})
\end{equation}
\begin{equation}
    F_{\text{part}}^{m} = \phi_{\text{part}}(X_{\text{rgb}}^{\text{part}}), \quad F_{\text{part}}^{i} = \phi_{\text{part}}(X_{\text{ir}}^{\text{part}})
\end{equation}

The extracted features are hierarchically fused to form a unified identity representation:
\begin{equation}
    H_{\text{fused}} = \phi_{\text{shared}}(F_{\text{global}}^{m} \oplus F_{\text{global}}^{i} \oplus F_{\text{part}}^{m} \oplus F_{\text{part}}^{i}),
\end{equation}

where \( \oplus \) denotes concatenation, and \( \phi_{\text{shared}} \) refines the concatenated features into a modality-invariant representation. Generalized Mean Pooling (GeM) then distills this representation for improved compactness and robustness.

\vspace{1em}  
\noindent\textbf{Loss Function.} HMG-DBNet utilizes identity loss, triplet loss, and MMD loss to achieve robust feature alignment and discrimination. The identity loss enhances separability across labels, while triplet loss optimizes intra-class and inter-class distinctions. MMD loss further reduces distribution gaps between RGB and IR features, collectively boosting cross-modality re-identification accuracy.

\subsection{Interactive Feature Fusion Strategy (IFFS)}

The \textbf{IFFS} framework addresses the inherent discrepancies between RGB and IR modalities by synergistically combining intra-modality fusion and cross-modality alignment. By capturing identity cues at both global and local levels, this dual strategy effectively enhances feature representation, alignment, and robustness in cross-modal person re-identification tasks.

\indent\textbf{Intra-Modality Feature Fusion.} The goal of intra-modality fusion is to refine identity-specific representations within each modality (RGB and IR) by combining complementary global and part-based features. This process integrates full-body and half-body views, ensuring that each modality captures both broad structural attributes and detailed local cues effectively. Formally, intra-modality fusion can be expressed as:
\begin{equation}
    F_{\text{intra}}^{\text{RGB}} = F_{\text{global}}^{\text{RGB}} \oplus F_{\text{part}}^{\text{RGB}}, \quad F_{\text{intra}}^{\text{IR}} = F_{\text{global}}^{\text{IR}} \oplus F_{\text{part}}^{\text{IR}},
\end{equation}
where \( \oplus \) denotes the concatenation operation. By merging expansive global features (e.g., body shape, overall appearance) with fine-grained local details (e.g., texture, distinctive identity markers), this fusion enhances the discriminative power of each modality. Specifically, it helps RGB features retain crucial spatial patterns that may be influenced by varying illumination, while IR features leverage unique thermal signatures to maintain robust identity representation under challenging environmental conditions.

\indent\textbf{Cross-Modality Feature Fusion.} While intra-modality fusion strengthens individual modality features, cross-modality fusion focuses on bridging the inherent distribution gap between RGB and IR data. Traditional fusion methods typically align full-body RGB features with full-body IR features, which often leads to loss of critical local information unique to each modality. To address this issue, we propose a novel fusion strategy that combines global features from full-body RGB images with localized features from half-body IR images, and vice versa:
\begin{equation}
    F_{\text{cross}}^{\text{RGB-IR}} = F_{\text{global}}^{\text{RGB}} \oplus F_{\text{part}}^{\text{IR}}, \quad F_{\text{cross}}^{\text{IR-RGB}} = F_{\text{global}}^{\text{IR}} \oplus F_{\text{part}}^{\text{RGB}}.
\end{equation}
\indent This cross-modality fusion leverages the complementary nature of RGB and IR features: full-body RGB features provide comprehensive structural information such as body posture and overall silhouette, while half-body IR features offer enhanced local details (e.g., facial contours, shoulder outlines) due to their sensitivity to thermal variations. By effectively aligning complementary characteristics from both modalities, this approach reduces the feature distribution gap, enhances feature alignment, and improves the model’s generalization capability. The result is a robust identity representation that performs well across diverse cross-modal re-identification scenarios.

\subsection{Adaptive Multi-Scale Kernel Maximum Mean Discrepancy (AMK-MMD)}

The AMK-MMD is designed to address complex feature distribution variations between RGB and IR modalities by leveraging multi-scale kernel fusion, adaptive bandwidth selection, and learnable kernel weights. Unlike traditional MMD, which typically employs a single kernel, AMK-MMD integrates multiple Gaussian kernels, each with distinct bandwidths, providing the flexibility to capture both coarse and fine-grained discrepancies across modalities.

The multi-scale kernel fusion, which captures diverse feature granularity, is defined as:
\vspace{-0.5em}
\begin{equation}
    K(\mathbf{x}, \mathbf{y}) = \sum_{m=1}^{M} \alpha_m \exp\left( -\frac{\| \mathbf{x} - \mathbf{y} \|_2^2}{2 \sigma_m^2} \right),
\end{equation}

where \smash{\(\{\sigma_m\}_{m=1}^{M}\)}  are bandwidths and \( \alpha_m \) are learnable weights that dynamically adjust during training. This adaptability enables AMK-MMD to capture essential distributional nuances, emphasizing the kernel scales that contribute most to feature alignment across heterogeneous data.

To improve computational efficiency, AMK-MMD leverages batch processing, symmetry, and vectorized computation, optimizing resource usage for larger datasets. The pairwise squared Euclidean distance matrix \( \mathbf{D} \) for features from source and target domains, combined into a single matrix \( \mathbf{Z} \), is calculated as:
\begin{equation}
    \mathbf{D}_{ij} = \| \mathbf{Z}_i - \mathbf{Z}_j \|_2^2.
\end{equation}
This formulation minimizes redundant calculations, facilitating scalability even for high-dimensional embeddings.

AMK-MMD further takes advantage of matrix symmetry, which significantly reduces computational load. Given the symmetrical property of the kernel matrix \( \mathbf{K} \) (i.e., \( \mathbf{K}_{ij} = \mathbf{K}_{ji} \)), only the upper triangular part of \( \mathbf{K} \) is computed, effectively halving the computational cost.

The AMK-MMD discrepancy loss, which quantifies feature alignment between modalities, is computed as:
\begin{equation}
\begin{aligned}
    \text{AMK-MMD}^2 = &\frac{1}{n_s(n_s - 1)} \sum_{\substack{i,j=1 \\ i \ne j}}^{n_s} \mathbf{K}_{ss}(i,j) \\
    &+ \frac{1}{n_t(n_t - 1)} \sum_{\substack{i,j=1 \\ i \ne j}}^{n_t} \mathbf{K}_{tt}(i,j) \\
    &- \frac{2}{n_s n_t} \sum_{i=1}^{n_s} \sum_{j=1}^{n_t} \mathbf{K}_{st}(i,j),
\end{aligned}
\end{equation}

where \( \mathbf{K}_{ss} \), \( \mathbf{K}_{tt} \), and \( \mathbf{K}_{st} \) are kernel submatrices representing source-source, target-target, and source-target comparisons. By leveraging batch processing and symmetry, AMK-MMD scales efficiently with large datasets while maintaining high alignment precision across modalities.

These features allow AMK-MMD to dynamically adapt to varying data distributions, effectively enhancing cross-modality re-identification through a balanced and computationally efficient approach to feature alignment.

\begin{table*}[htbp]
\footnotesize 

\centering 
\renewcommand{\arraystretch}{1.2} 
\setlength{\tabcolsep}{5pt} 

\begin{tabular}{cccccccccccccccccc}
\hline
                                    & \multicolumn{4}{c}{}                                                                   & \multicolumn{6}{c}{{SYSU-MM01}}                                                                                                                                          & \multicolumn{6}{c}{{RegDB}}                                                                                                                         \\ \hline
                                    & \multicolumn{4}{c|}{{Components}}                                               & \multicolumn{3}{c}{{All Search}}                                                & \multicolumn{3}{c|}{{Indoor Search}}                                            & \multicolumn{3}{c}{{Thermal to Visible}}                                        & \multicolumn{3}{c}{{Visible to Thermal}}                   \\ \hline
\multicolumn{1}{c|}{{Index}} & {Base} & {UBF} & {IMDAL} & \multicolumn{1}{c|}{{IDAL}} & {R-1} & {mAP} & \multicolumn{1}{c|}{{mINP}} & {R-1} & {mAP} & \multicolumn{1}{c|}{{mINP}} & {R-1} & {mAP} & \multicolumn{1}{c|}{{mINP}} & {R-1} & {mAP} & {mINP} \\ \hline
\multicolumn{1}{c|}{M0}             & \ding{51}                 &              &                & \multicolumn{1}{c|}{}              & 66.82            & 62.61        & \multicolumn{1}{c|}{47.90}         & 70.34            & 75.52        & \multicolumn{1}{c|}{70.86}         & 78.41            & 73.41        & \multicolumn{1}{c|}{61.36}         & 80.78            & 75.74        & 62.53         \\ \hline
\multicolumn{1}{c|}{M1}             & \ding{51}                 & \ding{51}            &                & \multicolumn{1}{c|}{}              & 71.65            & 64.97        & \multicolumn{1}{c|}{49.58}         & 75.87            & 78.49        & \multicolumn{1}{c|}{73.47}         & 83.77            & 76.92        & \multicolumn{1}{c|}{64.05}         & 84.86            & 78.95        & 65.33         \\ \hline
\multicolumn{1}{c|}{M2}             & \ding{51}                 & \ding{51}            & \ding{51}              & \multicolumn{1}{c|}{}              & 72.62            & 67.70        & \multicolumn{1}{c|}{51.19}         & 77.94            & 80.03        & \multicolumn{1}{c|}{75.72}         & 87.02            & 79.86        & \multicolumn{1}{c|}{66.65}         & 87.89            & 81.74        & 67.46         \\ \hline
\multicolumn{1}{c|}{M3}             & \ding{51}                 & \ding{51}            &                & \multicolumn{1}{c|}{\ding{51}}             & 72.23            & 67.04        & \multicolumn{1}{c|}{51.52}         & 77.35            & 79.55        & \multicolumn{1}{c|}{76.15}         & 87.69            & 80.04        & \multicolumn{1}{c|}{66.79}         & 88.37            & 82.57        & 67.70         \\ \hline
\multicolumn{1}{c|}{M4}             & \ding{51}                 & \ding{51}            & \ding{51}              & \multicolumn{1}{c|}{\ding{51}}             & 74.75            & 69.71        & \multicolumn{1}{c|}{53.32}         & 79.18            & 82.39        & \multicolumn{1}{c|}{77.43}         & 89.51            & 82.41        & \multicolumn{1}{c|}{68.04}         & 91.29            & 84.69        & 69.96         \\ \hline

\bottomrule
\end{tabular}
\caption{Ablation study results showing the impact of different components (UBF, IMDAL, and IDAL) on Rank-1 accuracy (R-1), mean Average Precision (mAP), and mean Inverse Negative Penalty (mINP) across SYSU-MM01 and RegDB datasets.}
\label{tab:comparison}
\end{table*}

\subsection{Phase-Enhanced Structural Attention Module (PESAM)
}

The \textbf{PESAM} leverages the unique properties of phase congruency to extract consistent, modality-invariant features across RGB and IR images, focusing on essential structural details such as edges and contours that remain unaffected by changes in lighting and contrast. This approach is particularly advantageous in cross-modality person re-identification, as phase congruency enables the network to identify structural features reliably across both RGB and IR modalities, providing a stable basis for alignment.

\begin{table}[]
\scriptsize 
\renewcommand{\arraystretch}{1.2} 
\setlength{\tabcolsep}{6pt} 
\centering
\begin{tabular}{cc|cccccc}
\hline
\multicolumn{2}{c|}{Weights} & \multicolumn{3}{c}{SYSU}                                         & \multicolumn{3}{c}{RegDB}                        \\ \hline
Intra         & Inter        & Rank-1         & mAP            & \multicolumn{1}{c|}{mINP}           & Rank-1         & mAP            & mINP           \\ \hline
0.0           & 1.0          & 71.05          & 65.15          & \multicolumn{1}{c|}{49.15}          & 89.13          & 81.46          & 66.80          \\
0.2           & 0.8          & 73.94          & 68.29          & \multicolumn{1}{c|}{52.87}          & \textbf{91.29} & \textbf{83.49} & \textbf{68.69} \\
0.4           & 0.6          & \textbf{74.75} & \textbf{69.71} & \multicolumn{1}{c|}{\textbf{53.32}} & 89.66          & 81.89          & 67.30          \\
0.6           & 0.4          & 73.07          & 67.01          & \multicolumn{1}{c|}{50.77}          & 87.52          & 81.18          & 66.75          \\
0.8           & 0.2          & 71.47          & 65.35          & \multicolumn{1}{c|}{49.44}          & 87.33          & 81.61          & 68.58          \\
1.0           & 0.0          & 71.60          & 65.16          & \multicolumn{1}{c|}{48.80}          & 88.16          & 81.80          & 67.38          \\ \hline
\end{tabular}
\caption{Results with different Intra and Inter weights on SYSU and RegDB datasets.}
\label{tab:intra weight}
\end{table}

\begin{figure}[h]
    \centering
    \begin{subfigure}{0.495\columnwidth} 
        \centering
        \includegraphics[width=\linewidth]{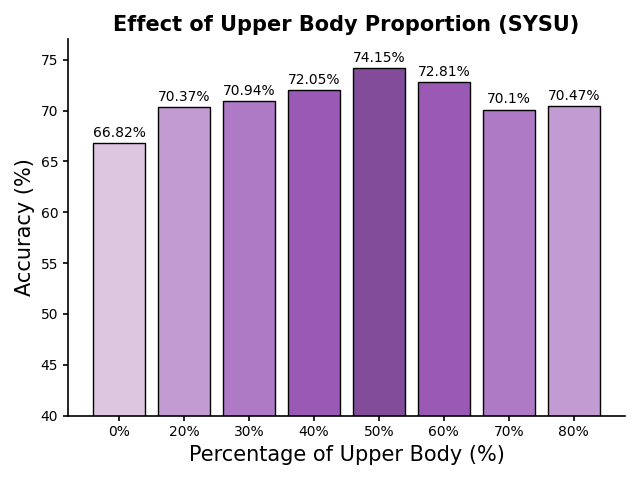} 
        \caption{Effect on SYSU}
        \label{fig:upper_body_proportion_sysu}
    \end{subfigure}
    \hfill
    \begin{subfigure}{0.495\columnwidth} 
        \centering
        \includegraphics[width=\linewidth]{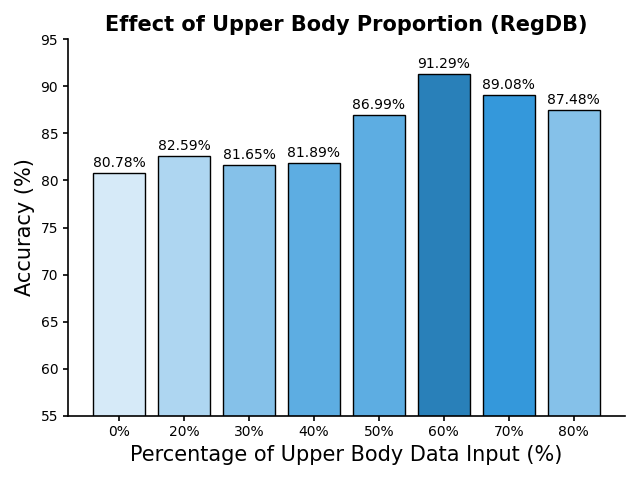} 
        \caption{Effect on RegDB}
        \label{fig:upper_body_proportion_regdb}
    \end{subfigure}

    \caption{Impact of Upper Body Proportion (UBP) on Model Accuracy for SYSU (left) and RegDB (right) datasets.}
    \label{fig:UBP}
\end{figure}

To compute the phase congruency map, we use the following formula:
\begin{equation}
    PC(x, y) = \frac{\sum_{s=1}^{S} \left( E_s(x, y) - T \right)}{\sum_{s=1}^{S} A_s(x, y) + \epsilon},
\end{equation}
where \( E_s(x, y) \) represents the phase congruency energy at scale \( s \), \( T \) is a noise threshold to filter out low-amplitude responses, and \( \epsilon \) prevents division by zero. This phase congruency map \( PC(x, y) \) highlights illumination-invariant edges and contours, allowing it to serve as a robust feature foundation across different lighting conditions.

\textbf{Edge-Guided Attention Mechanism (EGAM)} further enhances the model’s focus on modality-invariant features by generating an attention map from the phase congruency output. This attention map \( A(x, y) \), computed as:
\begin{equation}
    A(x, y) = \sigma(W_e * PC(x, y)),
\end{equation}
where \( \sigma(\cdot) \) is the sigmoid activation function, directs the network to concentrate on structurally relevant regions that are robust across RGB and IR domains. The EGAM effectively guides feature extraction, strengthening phase congruency’s role in aligning features from both modalities.

Then, we employ an adaptive weighting scheme to balance the contributions of RGB, IR, and phase congruency features in the final fused representation. This scheme dynamically adjusts each modality’s contribution to ensure balanced and robust feature fusion:
\vspace{-5pt}
\begin{equation}
    F_{\text{final}} = \alpha_{\text{vis}} F'_{\text{vis}} + \alpha_{\text{ir}} F'_{\text{ir}} + \alpha_{\text{phase}} F_{\text{phase}},
\end{equation}

where \( \alpha_{\text{vis}} \), \( \alpha_{\text{ir}} \), and \( \alpha_{\text{phase}} \) are learned weights for each modality.

%% file: 4_experiments.tex
\section{Experiments}

\subsection{Experimental Setups}

\noindent\textbf{Evaluation Metrics and Datasets.} We evaluate cross-modality Re-ID using three metrics: CMC, mAP, and mINP, to measure retrieval accuracy and robustness on two standard VI-ReID datasets:

\textbf{SYSU-MM01:} A large-scale dataset with 491 identities from six cameras (four RGB, two IR). The training set includes 395 identities; testing uses 96 identities with 3,803 IR queries and a gallery of 301 (single-shot) or 3,010 (multi-shot) RGB images, evaluated in all-search and indoor-search modes.

\textbf{RegDB:} Consists of 412 identities, each with 10 RGB and 10 IR images from one RGB and one thermal camera. Evaluations cover Thermal-to-Visible and Visible-to-Thermal modes, providing a comprehensive bidirectional assessment.

\begin{table}[htbp]
\scriptsize 
\centering 
\renewcommand{\arraystretch}{1.2} 
\setlength{\tabcolsep}{4pt} 

\begin{tabular}{l|c|ccc|ccc}
\hline
{}                            & {}                   & \multicolumn{3}{c|}{Thermal to Visible}                                                                                                                      & \multicolumn{3}{c}{Visible to Thermal}                                                                                                 \\ \hline
\multicolumn{1}{c|}{Method} & Venue              & R-1         & R-10        & mAP        & R-1         & R-10        & mAP            \\ \hline
Zero-Pad {[}35{]}           & ICCV 17            & 16.63          & 34.68          & 17.82          & 17.75          & 34.21          & 18.90          \\
eDBTR {[}42{]}              & TIFS 20            & 34.21          & 58.74          & 32.49          & 34.62          & 58.96          & 33.46          \\
D2RL {[}33{]}               & CVPR 19            & 43.40          & 66.10          & 44.10          & 43.40          & 66.10          & 44.10          \\

AlignGAN {[}30{]}           & ICCV 19            & 56.30          & -              & 53.40          & 57.90          & -              & 53.60          \\ \hline
DDAG {[}44{]}               & ECCV 20            & 68.08          & 85.15          & 61.80          & 69.34          & 86.19          & 63.46          \\
cm-SSFT {[}21{]}            & CVPR 20            & 71.00          & -              & 71.70          & 72.30          & -              & 72.90          \\ 
IMT {[}38{]}                & Neuro 21           & 56.30          & 71.33          & 88.11          & 75.49          & 87.48          & 69.64          \\
FBP-AL {[}34{]}             & TNNLS 21           & 70.05          & 89.22          & 66.61          & 73.98          & 89.71          & 68.24          \\
VSD {[}27{]}                & CVPR 21            & 71.80          & -              & 70.10          & 73.20          & -              & 71.60          \\

SMCL {[}59{]}               & ICCV 21            & 83.05          & -              & 78.57          & 83.93          & -              & 79.83          \\
MPANet {[}8{]}              & CVPR 21            & 83.70          & -              & 80.90          & 82.80          & -              & 80.70          \\ \hline

SPOT {[}2{]}                & TIP 22             & 79.37          & 92.79          & 72.26          & 80.35          & 93.48          & 72.46          \\
MAUM G {[}22{]}             & CVPR 22            & 81.07          & -              & 78.89          & 83.39          & -              & 78.75          \\
DART {[}33{]}               & CVPR 22            & 81.97          & -              & 73.78          & 83.60          & -              & 75.67          \\
SCFNet {[}24{]}             & ACCV 22            & 86.33          & 99.41          & 82.10          & 85.97          & 99.80          & 81.91          \\
TSME {[}34{]}               & TCSVT 22           & 86.41          & 96.39          & 75.70          & 87.35          & 97.10          & 76.94          \\

GDA {[}27{]}                & PR 23              & 69.67          & 86.41          & 61.98          & 73.95          & 89.47          & 65.49          \\
TOPLight {[}39{]}           & CVPR 23            & 80.65          & 92.81          & 75.91          & 85.51          & 94.99          & 79.95          \\
CMTR {[}63{]}               & TMM 23             & 81.06          & -              & 83.75          & 80.62          & -              & 74.42          \\
MTMFE {[}11{]}              & PR 23              & 81.11          & 92.35          & 79.59          & 85.04          & 94.38          & 82.52          \\
 \hline
\textbf{AMINet (Ours)}        & \textbf{This work} & \textbf{89.51} & \textbf{97.48} & \textbf{82.41} & \textbf{91.29} & \textbf{97.88} & \textbf{84.69} \\ \hline
\bottomrule
\end{tabular}
\caption{Performance comparisons on the RegDB dataset in both Thermal to Visible and Visible to Thermal modes.}
\label{tab:SOTA Regdb}
\end{table}

\noindent\textbf{Implementation Details.} Our method, implemented in PyTorch on a 24 GB RTX4090 GPU, resizes input images to 388×144 (global) and 194×144 (head-shoulder). Data augmentations include Random Cropping and Random Erasing. The model, based on ResNet-50 with Non-local blocks, is trained for 80 epochs with a batch size of 64. Feature extraction combines global and part-based modules aligned via a Contrastive Modal Aligner. GEM pooling and Batch Normalization stabilize training, while feature mixing enhances cross-modal discrimination. Optimization uses SGD with weight decay \(5 \times 10^{-4}\), momentum 0.9, and a staged learning rate starting at 0.01, peaking at 0.1, and reducing to 0.001.

\vspace{1em}  
\subsection{Ablation Study}

As shown in \cref{tab:comparison}, to evaluate the impact of each module in our model on cross-modality person re-identification (Re-ID) performance, we conducted a series of ablation experiments involving three key components: Upper Body Feature extraction (UBF), Intra-Modality Distribution Alignment Loss (IMDAL), and Inter-Modality Distribution Alignment Loss (IDAL).

\vspace{1em}  
\noindent\textbf{Baseline Model.}  
The baseline model (M0), which only includes full-body feature extraction (Base) without any additional modules, achieved 66.82\% Rank-1 accuracy in the SYSU-MM01 all-search mode and 70.34\% in indoor-search. On the RegDB dataset, it scored 78.41\% Rank-1 accuracy in the Thermal-to-Visible mode and 80.78\% in the Visible-to-Thermal mode. These results indicate that using whole-body features alone limits the model's ability to handle cross-modality discrepancies, particularly under occlusion and varying illumination.

\begin{figure}[htbp]
    \centering
    \begin{subfigure}{0.49\columnwidth}
        \centering
        \includegraphics[width=\linewidth]{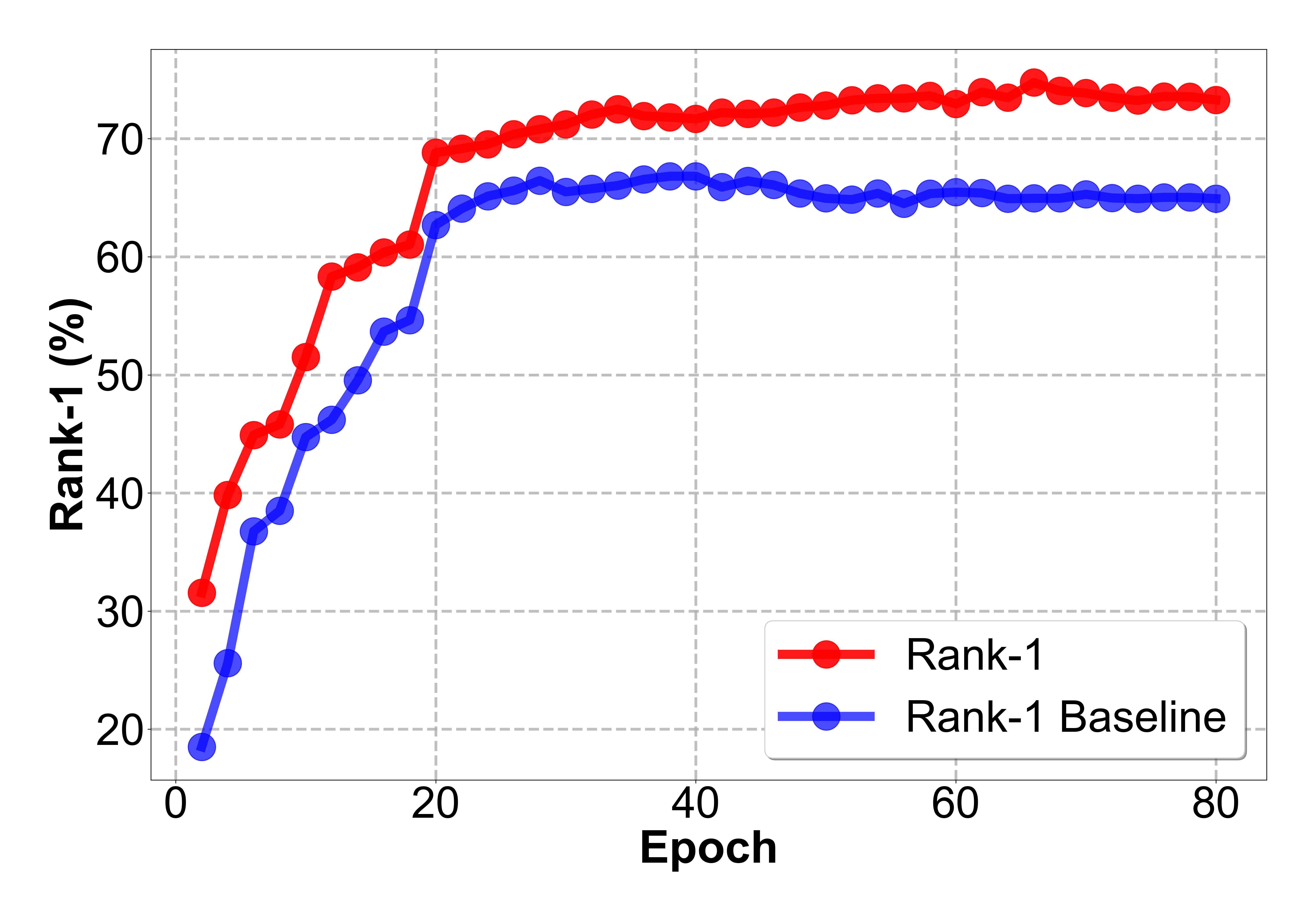}
        \caption{Rank-1 (SYSU)}
        \label{fig:rank1_sysu}
    \end{subfigure}
    \hfill
    \begin{subfigure}{0.49\columnwidth}
        \centering
        \includegraphics[width=\linewidth]{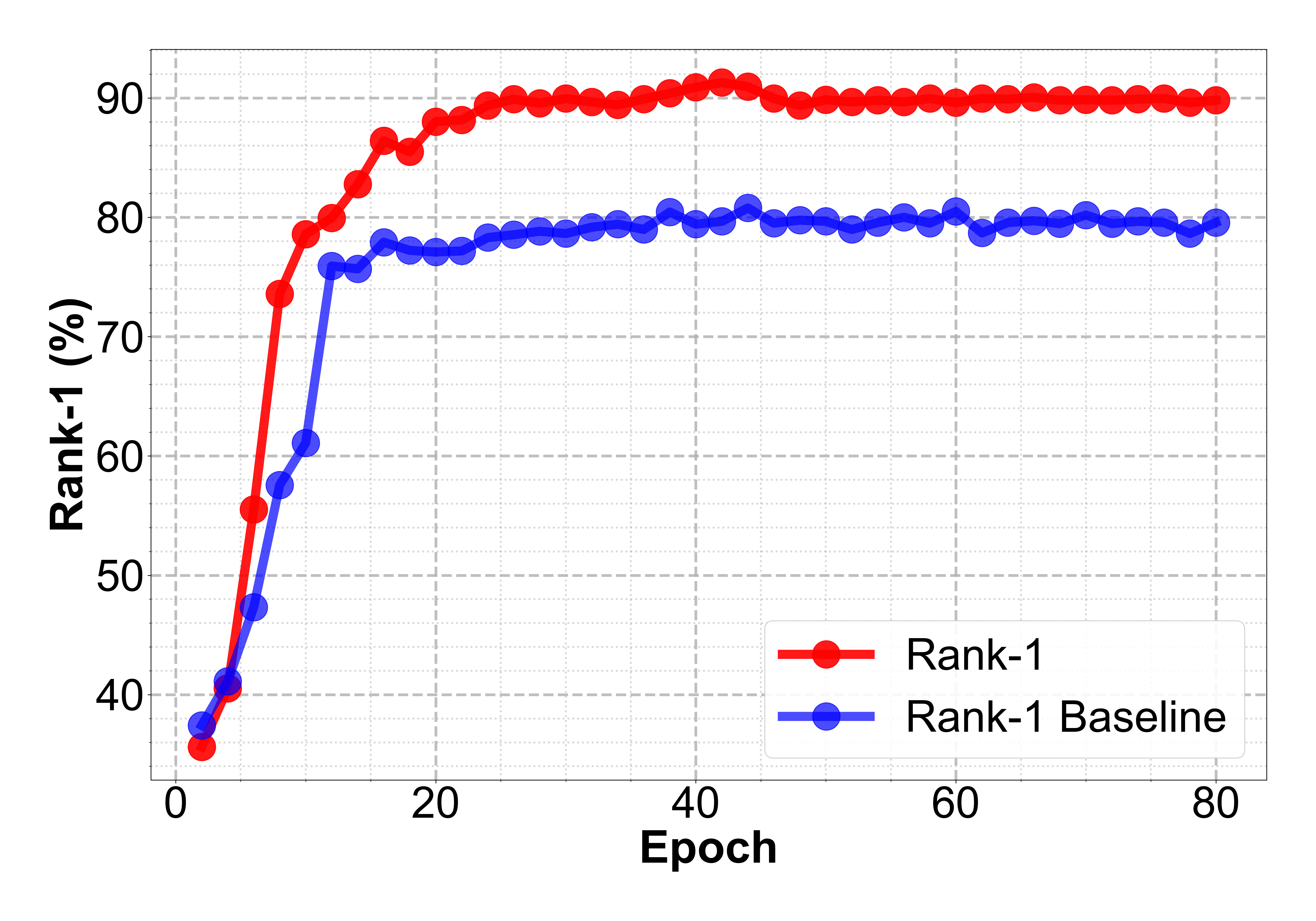}
        \caption{Rank-1 (RegDB)}
        \label{fig:rank1_regdb}
    \end{subfigure}

    \vspace{0.5em}

    \begin{subfigure}{0.49\columnwidth}
        \centering
        \includegraphics[width=\linewidth]{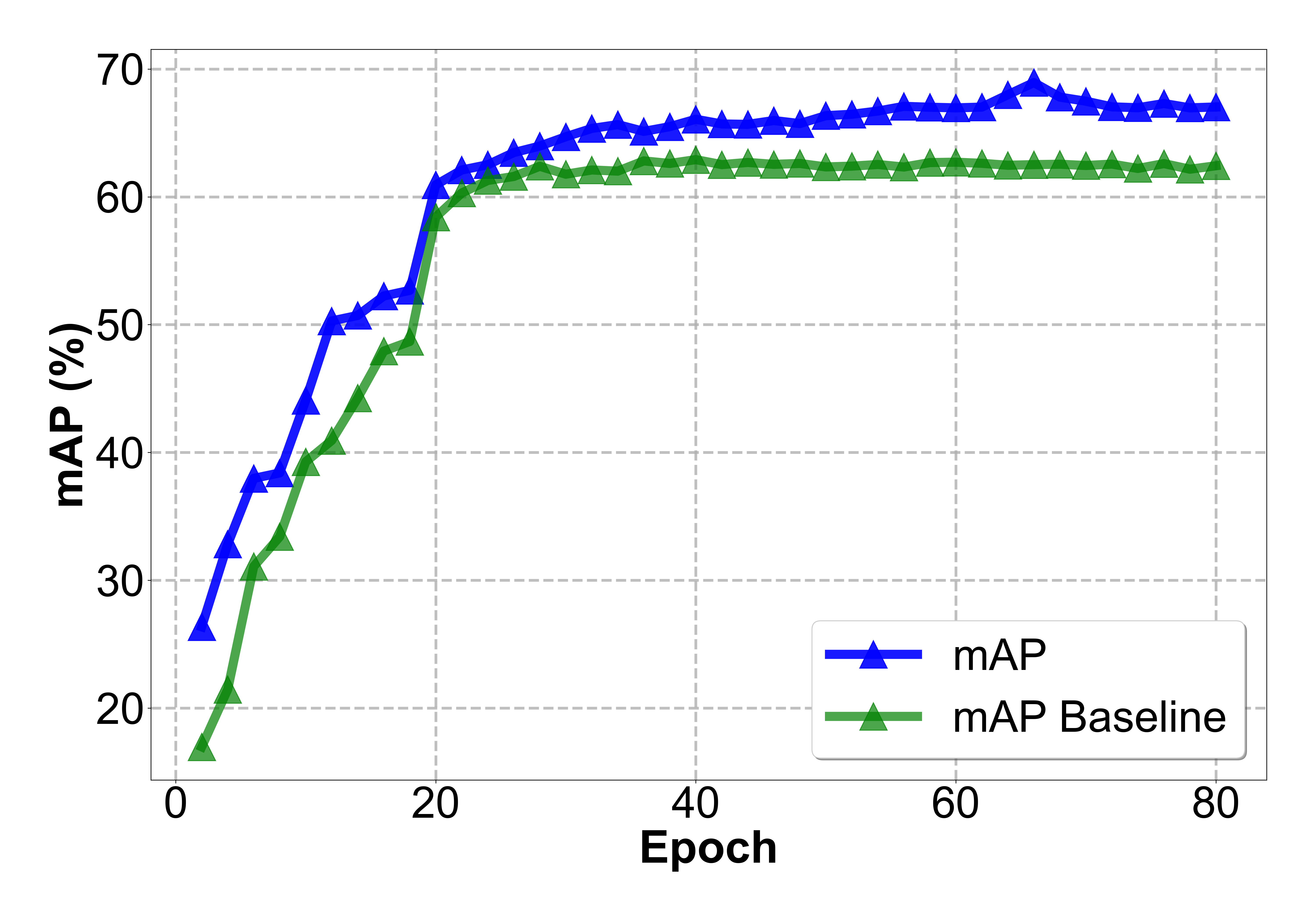}
        \caption{mAP (SYSU)}
        \label{fig:map_sysu}
    \end{subfigure}
    \hfill
    \begin{subfigure}{0.49\columnwidth}
        \centering
        \includegraphics[width=\linewidth]{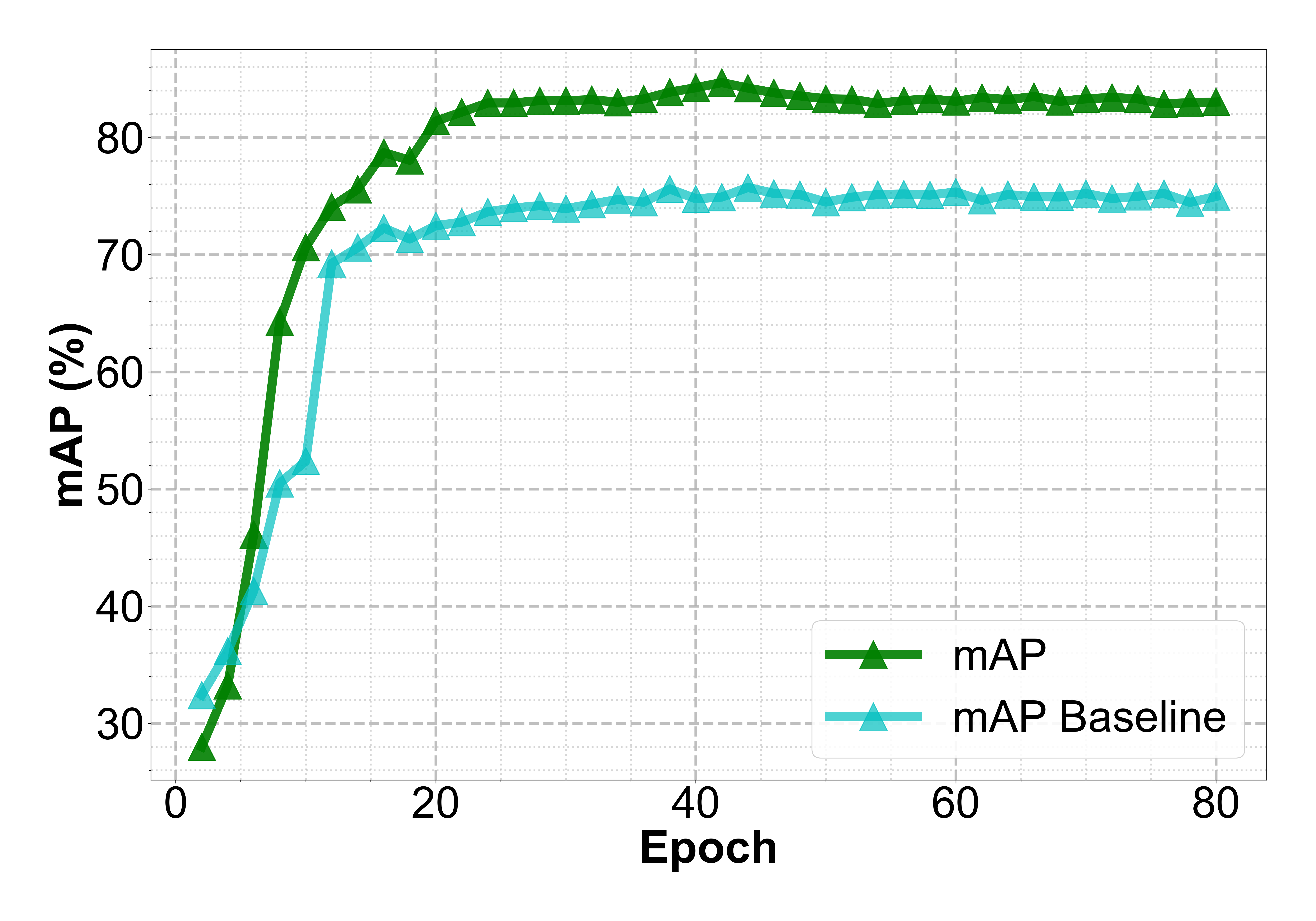}
        \caption{mAP (RegDB)}
        \label{fig:map_regdb}
    \end{subfigure}

    \vspace{0.5em} 
    \caption{Rank-1 and mAP results on SYSU and RegDB datasets.}

\end{figure}

\vspace{1em}  
\noindent\textbf{Upper Body Feature Extraction (UBF).}  
Introducing the UBF module (Model M1) improved the Rank-1 accuracy from 66.82\% to 71.65\% in SYSU-MM01 all-search and from 78.41\% to 83.77\% in RegDB Thermal-to-Visible. The UBF module captures fine-grained details from the upper body, such as head, shoulders, and chest, which enhances the model's robustness, particularly in cases with occlusion or pose variation. This module highlights the importance of local features for maintaining discriminative power across different modalities.

\vspace{1em}  
\noindent\textbf{Intra-Modality Distribution Alignment (IMDAL).}  
Building on M1, the addition of IMDAL (Model M2) further increased the Rank-1 accuracy to 72.62\% in SYSU-MM01 all-search and to 87.02\% in RegDB Thermal-to-Visible. IMDAL reduces intra-modality variability by aligning feature distributions within each modality, addressing inconsistencies caused by lighting, pose, and other variations. This enhanced intra-modality alignment improves feature stability and robustness, laying a solid foundation for effective cross-modality alignment.

\vspace{1em}  
\noindent\textbf{Inter-Modality Distribution Alignment (IDAL).}  
Adding IDAL to M1 (Model M3) improved the Rank-1 accuracy from 71.65\% to 72.23\% in SYSU-MM01 all-search and from 83.77\% to 87.69\% in RegDB Thermal-to-Visible. IDAL minimizes discrepancies between RGB and infrared feature distributions, effectively aligning them within a shared feature space. By reducing cross-modality matching errors, IDAL enables better generalization across modalities, capturing intrinsic correlations and enhancing model robustness under diverse modality conditions.


\begin{table*}[htbp]
\centering 
\footnotesize
\renewcommand{\arraystretch}{1.10} 
\setlength{\tabcolsep}{9pt} 


\begin{tabular}{l|c|cccccccccccc}
\hline

{\color[HTML]{000000} \textbf{}}                            & {\color[HTML]{000000} \textbf{}}      & \multicolumn{5}{c}{{\color[HTML]{000000} \textbf{All Search}}}                                                                                                                                                               & \multicolumn{5}{c}{{\color[HTML]{000000} \textbf{Indoor Search}}}                                                                                                                                       \\ \hline
\multicolumn{1}{c|}{{\color[HTML]{000000} \textbf{Method}}} & {\color[HTML]{000000} \textbf{Venue}} & {\color[HTML]{000000} \textbf{R-1}} & {\color[HTML]{000000} \textbf{R-10}} & {\color[HTML]{000000} \textbf{R-20}} & {\color[HTML]{000000} \textbf{mAP}} & \multicolumn{1}{c|}{{\color[HTML]{000000} \textbf{mINP}}} & {\color[HTML]{000000} \textbf{R-1}} & {\color[HTML]{000000} \textbf{R-10}} & {\color[HTML]{000000} \textbf{R-20}} & {\color[HTML]{000000} \textbf{mAP}} & {\color[HTML]{000000} \textbf{mINP}} \\ \hline
{\color[HTML]{000000} eDBTR {[}42{]}}                       & {\color[HTML]{000000} TIFS 20}        & {\color[HTML]{000000} 27.82}           & {\color[HTML]{000000} 67.34}            & {\color[HTML]{000000} 81.34}            & {\color[HTML]{000000} 28.42}        & \multicolumn{1}{c|}{{\color[HTML]{000000} \textbf{-}}}    & {\color[HTML]{000000} 32.46}           & {\color[HTML]{000000} 77.42}            & {\color[HTML]{000000} 89.62}            & {\color[HTML]{000000} 42.46}        & {\color[HTML]{000000} \textbf{-}}    \\
{\color[HTML]{000000} CoSiGAN {[}56{]}}                     & {\color[HTML]{000000} ICMR 20}        & {\color[HTML]{000000} 35.55}           & {\color[HTML]{000000} 81.54}            & {\color[HTML]{000000} 90.43}            & {\color[HTML]{000000} 38.33}        & \multicolumn{1}{c|}{{\color[HTML]{000000} \textbf{-}}}    & {\color[HTML]{000000} \textbf{-}}      & {\color[HTML]{000000} \textbf{-}}       & {\color[HTML]{000000} \textbf{-}}       & {\color[HTML]{000000} \textbf{-}}   & {\color[HTML]{000000} \textbf{-}}    \\
{\color[HTML]{000000} MSR {[}5{]}}                          & {\color[HTML]{000000} TIP 20}         & {\color[HTML]{000000} 37.35}           & {\color[HTML]{000000} 83.40}            & {\color[HTML]{000000} 93.34}            & {\color[HTML]{000000} 38.11}        & \multicolumn{1}{c|}{{\color[HTML]{000000} \textbf{-}}}    & {\color[HTML]{000000} 39.64}           & {\color[HTML]{000000} 89.29}            & {\color[HTML]{000000} 97.66}            & {\color[HTML]{000000} 50.88}        & {\color[HTML]{000000} \textbf{-}}    \\
{\color[HTML]{000000} X-Modal {[}12{]}}                     & {\color[HTML]{000000} AAAI 20}        & {\color[HTML]{000000} 49.92}           & {\color[HTML]{000000} 89.79}            & {\color[HTML]{000000} 95.96}            & {\color[HTML]{000000} 50.73}        & \multicolumn{1}{c|}{{\color[HTML]{000000} \textbf{-}}}    & {\color[HTML]{000000} \textbf{-}}      & {\color[HTML]{000000} \textbf{-}}       & {\color[HTML]{000000} \textbf{-}}       & {\color[HTML]{000000} \textbf{-}}   & {\color[HTML]{000000} \textbf{-}}    \\
{\color[HTML]{000000} DDAG {[}44{]}}                        & {\color[HTML]{000000} ECCV 20}        & {\color[HTML]{000000} 54.75}           & {\color[HTML]{000000} 90.36}            & {\color[HTML]{000000} 95.81}            & {\color[HTML]{000000} 53.02}        & \multicolumn{1}{c|}{{\color[HTML]{000000} 39.62}}         & {\color[HTML]{000000} 61.02}           & {\color[HTML]{000000} 94.06}            & {\color[HTML]{000000} 98.41}            & {\color[HTML]{000000} 67.98}        & {\color[HTML]{000000} 62.61}         \\
{\color[HTML]{000000} LLM {[}4{]}}                          & {\color[HTML]{000000} ECCV 20}        & {\color[HTML]{000000} 55.25}           & {\color[HTML]{000000} 86.09}            & {\color[HTML]{000000} 92.69}            & {\color[HTML]{000000} 52.96}        & \multicolumn{1}{c|}{{\color[HTML]{000000} \textbf{-}}}    & {\color[HTML]{000000} 59.65}           & {\color[HTML]{000000} 90.85}            & {\color[HTML]{000000} 95.02}            & {\color[HTML]{000000} 65.46}        & {\color[HTML]{000000} \textbf{-}}    \\
 \hline

{\color[HTML]{000000} IMT {[}38{]}}                         & {\color[HTML]{000000} Neuro 21}       & {\color[HTML]{000000} 56.52}           & {\color[HTML]{000000} 90.26}            & {\color[HTML]{000000} 95.59}            & {\color[HTML]{000000} 57.47}        & \multicolumn{1}{c|}{{\color[HTML]{000000} 38.75}}         & {\color[HTML]{000000} 68.72}           & {\color[HTML]{000000} 94.61}            & {\color[HTML]{000000} 97.42}            & {\color[HTML]{000000} 75.22}        & {\color[HTML]{000000} 64.22}         \\
{\color[HTML]{000000} NFS {[}1{]}}                          & {\color[HTML]{000000} CVPR 21}        & {\color[HTML]{000000} 56.91}           & {\color[HTML]{000000} 91.34}            & {\color[HTML]{000000} 96.52}            & {\color[HTML]{000000} 55.45}        & \multicolumn{1}{c|}{{\color[HTML]{000000} \textbf{-}}}    & {\color[HTML]{000000} 62.79}           & {\color[HTML]{000000} 96.53}            & {\color[HTML]{000000} 99.07}            & {\color[HTML]{000000} 69.79}        & {\color[HTML]{000000} \textbf{-}}    \\
{\color[HTML]{000000} VSD {[}27{]}}                         & {\color[HTML]{000000} CVPR 21}        & {\color[HTML]{000000} 60.02}           & {\color[HTML]{000000} 94.18}            & {\color[HTML]{000000} 98.14}            & {\color[HTML]{000000} 58.80}        & \multicolumn{1}{c|}{{\color[HTML]{000000} \textbf{-}}}    & {\color[HTML]{000000} 66.05}           & {\color[HTML]{000000} 96.59}            & {\color[HTML]{000000} 99.38}            & {\color[HTML]{000000} 72.98}        & {\color[HTML]{000000} \textbf{-}}    \\
{\color[HTML]{000000} GLMC {[}48{]}}                        & {\color[HTML]{000000} TNNLS 21}       & {\color[HTML]{000000} 64.37}           & {\color[HTML]{000000} 93.90}            & {\color[HTML]{000000} 97.53}            & {\color[HTML]{000000} 63.43}        & \multicolumn{1}{c|}{{\color[HTML]{000000} \textbf{-}}}    & {\color[HTML]{000000} 67.35}           & {\color[HTML]{000000} 98.10}            & {\color[HTML]{000000} 99.77}            & {\color[HTML]{000000} 74.02}        & {\color[HTML]{000000} \textbf{-}}    \\

{\color[HTML]{000000} MCLNet {[}8{]}}                       & {\color[HTML]{000000} ICCV 21}        & {\color[HTML]{000000} 65.40}           & {\color[HTML]{000000} 93.33}            & {\color[HTML]{000000} 97.14}            & {\color[HTML]{000000} 61.98}        & \multicolumn{1}{c|}{{\color[HTML]{000000} 47.39}}         & {\color[HTML]{000000} 72.56}           & {\color[HTML]{000000} 96.98}            & {\color[HTML]{000000} 99.20}            & {\color[HTML]{000000} 76.58}        & {\color[HTML]{000000} 72.10}         \\
{\color[HTML]{000000} SMCL {[}59{]}}                        & {\color[HTML]{000000} ICCV 21}        & {\color[HTML]{000000} 67.39}           & {\color[HTML]{000000} 92.87}            & {\color[HTML]{000000} 96.76}            & {\color[HTML]{000000} 61.78}        & \multicolumn{1}{c|}{{\color[HTML]{000000} \textbf{-}}}    & {\color[HTML]{000000} 68.84}           & {\color[HTML]{000000} 96.55}            & {\color[HTML]{000000} 98.77}            & {\color[HTML]{000000} 75.56}        & {\color[HTML]{000000} \textbf{-}}   
 \\ \hline
{\color[HTML]{000000} DML {[}42{]}}                         & {\color[HTML]{000000} TCSVT 22}       & {\color[HTML]{000000} 58.40}           & {\color[HTML]{000000} 91.20}            & {\color[HTML]{000000} 96.90}            & {\color[HTML]{000000} 56.10}        & \multicolumn{1}{c|}{{\color[HTML]{000000} -}}             & {\color[HTML]{000000} 62.40}           & {\color[HTML]{000000} 95.20}            & {\color[HTML]{000000} 98.70}            & {\color[HTML]{000000} 69.50}        & {\color[HTML]{000000} -}             \\
{\color[HTML]{000000} MAUM G {[}22{]}}                      & {\color[HTML]{000000} CVPR 22}        & {\color[HTML]{000000} 61.59}           & {\color[HTML]{000000} -}                & {\color[HTML]{000000} -}                & {\color[HTML]{000000} 59.96}        & \multicolumn{1}{c|}{{\color[HTML]{000000} -}}             & {\color[HTML]{000000} 67.07}           & {\color[HTML]{000000} -}                & {\color[HTML]{000000} -}                & {\color[HTML]{000000} 73.58}        & {\color[HTML]{000000} -}             \\
{\color[HTML]{000000} TSME {[}34{]}}                        & {\color[HTML]{000000} TCSVT 22}       & {\color[HTML]{000000} 64.23}           & {\color[HTML]{000000} 95.19}            & {\color[HTML]{000000} 98.73}            & {\color[HTML]{000000} 61.21}        & \multicolumn{1}{c|}{{\color[HTML]{000000} -}}             & {\color[HTML]{000000} 64.8}            & {\color[HTML]{000000} 96.92}            & {\color[HTML]{000000} 99.31}            & {\color[HTML]{000000} 71.53}        & {\color[HTML]{000000} -}             \\
{\color[HTML]{000000} SPOT {[}2{]}}                         & {\color[HTML]{000000} TIP 22}         & {\color[HTML]{000000} 65.34}           & {\color[HTML]{000000} 92.73}            & {\color[HTML]{000000} 97.04}            & {\color[HTML]{000000} 62.25}        & \multicolumn{1}{c|}{{\color[HTML]{000000} -}}             & {\color[HTML]{000000} 69.42}           & {\color[HTML]{000000} 96.22}            & {\color[HTML]{000000} 99.12}            & {\color[HTML]{000000} 74.63}        & {\color[HTML]{000000} -}             \\
{\color[HTML]{000000} FMCNet {[}43{]}}                      & {\color[HTML]{000000} CVPR 22}        & {\color[HTML]{000000} 66.34}           & {\color[HTML]{000000} -}                & {\color[HTML]{000000} -}                & {\color[HTML]{000000} 62.51}        & \multicolumn{1}{c|}{{\color[HTML]{000000} -}}             & {\color[HTML]{000000} 68.15}           & {\color[HTML]{000000} -}                & {\color[HTML]{000000} -}                & {\color[HTML]{000000} 74.09}        & {\color[HTML]{000000} -}             \\
{\color[HTML]{000000} DART {[}33{]}}                        & {\color[HTML]{000000} CVPR 22}        & {\color[HTML]{000000} 68.72}           & {\color[HTML]{000000} 96.39}            & {\color[HTML]{000000} 98.96}            & {\color[HTML]{000000} 66.29}        & \multicolumn{1}{c|}{{\color[HTML]{000000} -}}             & {\color[HTML]{000000} 72.52}           & {\color[HTML]{000000} 97.84}            & {\color[HTML]{000000} 99.46}            & {\color[HTML]{000000} 78.17}        & {\color[HTML]{000000} -}           
\\ \hline
{\color[HTML]{000000} GDA {[}27{]}}                         & {\color[HTML]{000000} PR 23}          & {\color[HTML]{000000} 63.94}           & {\color[HTML]{000000} 93.34}            & {\color[HTML]{000000} 97.29}            & {\color[HTML]{000000} 60.73}        & \multicolumn{1}{c|}{{\color[HTML]{000000} -}}             & {\color[HTML]{000000} 71.06}           & {\color[HTML]{000000} 97.31}            & {\color[HTML]{000000} 99.47}            & {\color[HTML]{000000} 76.01}        & {\color[HTML]{000000} -}             \\
{\color[HTML]{000000} CMTR {[}63{]}}                        & {\color[HTML]{000000} TMM 23}         & {\color[HTML]{000000} 65.45}           & {\color[HTML]{000000} 94.47}            & {\color[HTML]{000000} 98.16}            & {\color[HTML]{000000} 62.90}        & \multicolumn{1}{c|}{{\color[HTML]{000000} -}}             & {\color[HTML]{000000} 71.64}           & {\color[HTML]{000000} 97.16}            & {\color[HTML]{000000} 99.22}            & {\color[HTML]{000000} 76.67}        & {\color[HTML]{000000} -}             \\
{\color[HTML]{000000} TOPLight {[}39{]}}                    & {\color[HTML]{000000} CVPR 23}        & {\color[HTML]{000000} 66.76}           & {\color[HTML]{000000} 96.23}            & {\color[HTML]{000000} 98.70}            & {\color[HTML]{000000} 64.01}        & \multicolumn{1}{c|}{{\color[HTML]{000000} -}}             & {\color[HTML]{000000} 72.89}           & {\color[HTML]{000000} 97.93}            & {\color[HTML]{000000} 99.28}            & {\color[HTML]{000000} 76.70}        & {\color[HTML]{000000} -}             \\
{\color[HTML]{000000} MRCN {[}61{]}}                        & {\color[HTML]{000000} AAAI 23}        & {\color[HTML]{000000} 68.90}           & {\color[HTML]{000000} 95.20}            & {\color[HTML]{000000} 98.40}            & {\color[HTML]{000000} 65.50}        & \multicolumn{1}{c|}{{\color[HTML]{000000} -}}             & {\color[HTML]{000000} 76.00}           & {\color[HTML]{000000} 98.30}            & {\color[HTML]{000000} 99.70}            & {\color[HTML]{000000} 79.80}        & {\color[HTML]{000000} -}             \\
{\color[HTML]{000000} MTMFE {[}11{]}}                       & {\color[HTML]{000000} PR 23}          & {\color[HTML]{000000} 69.47}           & {\color[HTML]{000000} 96.42}            & {\color[HTML]{000000} 99.11}            & {\color[HTML]{000000} 66.41}        & \multicolumn{1}{c|}{{\color[HTML]{000000} -}}             & {\color[HTML]{000000} 71.72}           & {\color[HTML]{000000} 97.19}            & {\color[HTML]{000000} 98.97}            & {\color[HTML]{000000} 76.38}        & {\color[HTML]{000000} -}             \\
{\color[HTML]{000000} MRCN-P {[}61{]}}                      & {\color[HTML]{000000} AAAI 23}        & {\color[HTML]{000000} 70.80}           & {\color[HTML]{000000} 96.50}            & {\color[HTML]{000000} 99.10}            & {\color[HTML]{000000} 67.30}        & \multicolumn{1}{c|}{{\color[HTML]{000000} -}}             & {\color[HTML]{000000} 76.40}           & {\color[HTML]{000000} 98.50}            & {\color[HTML]{000000} 99.90}            & {\color[HTML]{000000} 80.00}        & {\color[HTML]{000000} -}            
 \\ \hline
\textbf{{\color[HTML]{000000} AMINet (Ours)}}                         & \textbf{{\color[HTML]{000000} This work}}      & \textbf{{\color[HTML]{000000} 74.75}}           & \textbf{{\color[HTML]{000000} 97.87}}            & \textbf{{\color[HTML]{000000} 99.32}}            & \textbf{{\color[HTML]{000000} 66.11}}        & \multicolumn{1}{c|}{\textbf{{\color[HTML]{000000} 51.32}}}         & \textbf{{\color[HTML]{000000} 79.18}}           & \textbf{{\color[HTML]{000000} 98.78}}            & \textbf{{\color[HTML]{000000} 99.67}}            & \textbf{{\color[HTML]{000000} 82.39}}        & \textbf{{\color[HTML]{000000} 77.43}}         \\ \hline

\bottomrule
\end{tabular}

\caption{Performance comparisons on SYSU-MM01 dataset in both All Search and Indoor Search modes. Our method, denoted as AMINet (Ours), significantly outperforms existing methods across key evaluation metrics, including Rank-1, Rank-10, Rank-20, mAP, and mINP.}
\label{tab:sysu SOTA}
\end{table*}

\subsection{Effect of Loss Weights}

As shown in \cref{tab:intra weight}, the results indicate clear differences in the optimal weights for intra-modality and inter-modality alignment under MMD constraints across datasets. This highlights the distinct modality characteristics of each dataset. In the SYSU, the RGB and IR images exhibit a certain degree of consistency in lighting conditions, viewpoints, and environmental changes, resulting in relatively smaller modality discrepancies. The best performance is achieved with an intra-modality weight of 0.4 and an inter-modality weight of 0.6. Conversely, the RegDB has greater modality differences, necessitating a higher inter-modality weight of 0.8. This indicates that stronger cross-modality alignment is required to bridge the larger distribution gap.

\subsection{Effect of Upper-Body Proportion}
As shown in \cref{fig:UBP}, experiments on SYSU and RegDB datasets assess the impact of upper-body data proportion on model performance in cross-modality person re-identification. Results show a strong correlation between upper-body proportion and accuracy. For SYSU, peak accuracy (74.15\%) occurs at 50\% upper-body input, while RegDB reaches its highest (91.29\%) at 60\%. This suggests an optimal range of 50\%-60\%, where key features like shoulders and torso enhance RGB-IR alignment. Notably, SYSU accuracy declines beyond 50\%, indicating redundancy from excessive upper-body data, which dilutes critical full-body cues. In contrast, RegDB shows stability at 60\%, benefiting from consistent scene conditions. These findings underscore the importance of a balanced upper-body ratio for effective feature alignment and robust identification.

\begin{figure*}[t]  
    \centering
    \includegraphics[width=\textwidth]{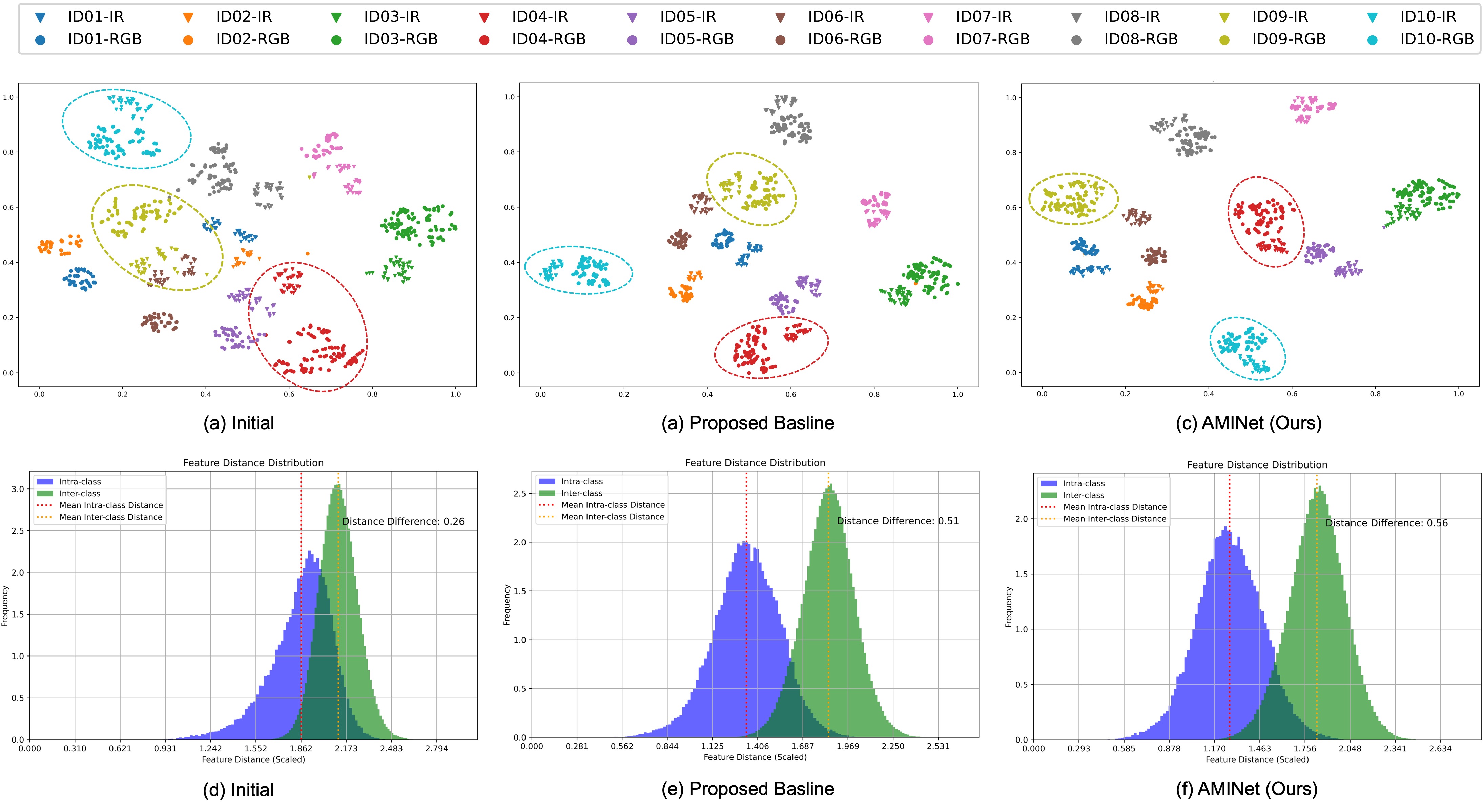}  
    \captionsetup{justification=justified, singlelinecheck=false}  
    \caption{t-SNE Visualizations and Feature Distance Distributions for Different Models, Demonstrating Cross-Modality Feature Alignment and Intra/Inter-Class Separation.}
    \label{fig:tsne_distance}
\end{figure*}

\subsection{State-of-the-Art Performance Comparison}

\noindent\textbf{Evaluation on SYSU-MM01.} Our method surpasses state-of-the-art models on the challenging SYSU-MM01 dataset, demonstrating superior performance in both All Search and Indoor Search modes, as shown in \cref{tab:sysu SOTA}. In All Search, it reaches a Rank-1 accuracy of 74.75\%, mAP of 66.11\%, and mINP of 51.32\%, surpassing SCFNet by 1.79\% in Rank-1 and 11.78\% in mINP. This superior alignment reflects the method's robustness in unifying RGB and IR features effectively. In Indoor Search, our model further demonstrates its adaptability under controlled settings, achieving 79.18\% in Rank-1 accuracy and 82.39\% in mAP, surpassing MPANet by 2.44\% and 1.44\%, respectively. These consistent gains highlight the method’s effectiveness in handling intra- and cross-modality challenges.

\noindent\textbf{Evaluation on RegDB.} On the RegDB dataset, which encompasses both Thermal-to-Visible and Visible-to-Thermal modes, our model once again achieves state-of-the-art results, as shown in \cref{tab:SOTA Regdb}. In Thermal-to-Visible mode, it records 89.51\% in Rank-1 accuracy and 82.41\% in mAP, improving upon MAUM P by 2.56\% and 3.07\%, respectively. In Visible-to-Thermal mode, it reaches a Rank-1 accuracy of 91.29\% and mAP of 84.69\%, with a 3.42\% gain in Rank-1 over MCLNet. These results reflect the model’s advanced feature alignment and fusion strategies, effectively bridging RGB and IR modalities to achieve high re-identification accuracy across varied conditions.

\subsection{Visual Analysis}

\noindent\textbf{t-SNE Visualization of Extracted Embeddings.} \cref{fig:tsne_distance} presents the t-SNE visualizations of feature embeddings from RGB and IR modalities. In Figure 4(a), the initial model shows dispersed identity clusters with clear separation between RGB (circles) and IR (triangles), indicating poor cross-modality alignment. Figure 4(b) illustrates our newly integrated baseline model, which achieves improved clustering but still exhibits inconsistent modality alignment. In contrast, Figure 4(c) shows proposed AMINet model, where identity clusters are distinct and tightly grouped, including both RGB and IR samples, demonstrating superior alignment and identity discrimination, and highlighting the model’s strong performance.

\noindent\textbf{Cosine Similarity Analysis.} Figures 4(d), 4(e), and 4(f) illustrate the intra- and inter-class feature distance distributions to assess model discrimination. The x-axis shows Euclidean distances, and the y-axis shows frequency. Blue lines represent intra-class (same identity), while green lines represent inter-class (different identities). In Figure 4(d), the initial model shows significant overlap with a small mean gap of 0.26, reflecting weak discrimination. Figure 4(e) shows the baseline model increasing this gap, improving separation. Our final model in Figure 4(f) further expands this gap to 0.56, with minimal overlap, indicating enhanced clustering of intra-class samples and clear separation of inter-class samples. This confirms the superior performance of our model in cross-modality feature alignment and identity discrimination.

%% file: 5_conclusion.tex
\section{Conclusion}

In this paper, we propose AMINet, an Adaptive Modality Interaction Network for Visible-Infrared Person Re-Identification (VI-ReID). AMINet integrates multi-granularity feature extraction (HMG-DBNet), interactive feature fusion (IFFS), phase-enhanced attention (PESAM), and adaptive alignment (AMK-MMD) to bridge the RGB-IR modality gap effectively. Extensive experiments on SYSU-MM01 and RegDB datasets validate our approach, achieving a new state-of-the-art Rank-1 accuracy of 74.75\% on SYSU-MM01 and {91.29\% on RegDB. These results demonstrate the robustness and generalization of our model, offering a strong baseline for future VI-ReID research.